\documentclass[letterpaper]{article} 
\usepackage{aaai23}  
\usepackage{times}  
\usepackage{helvet}  
\usepackage{courier}  
\usepackage[hyphens]{url}  
\usepackage{graphicx} 
\usepackage{subfig}
\usepackage{natbib}  
\usepackage{caption} 
\usepackage{algorithm}
\usepackage{algorithmic}
\usepackage{makecell}
\usepackage{newfloat}
\usepackage{listings}
\DeclareCaptionStyle{ruled}{labelfont=normalfont,labelsep=colon,strut=off} 
\floatstyle{ruled}
\newfloat{listing}{tb}{lst}{}
\floatname{listing}{Listing}
\title{Multi-Task Learning for Budbreak Prediction}
\author{
    Aseem Saxena\textsuperscript{\rm 1}, 
    Paola Pesantez-Cabrera\textsuperscript{\rm 2},
    Rohan Ballapragada\textsuperscript{\rm 1},
    Markus Keller,\textsuperscript{\rm 3} 
    Alan Fern\textsuperscript{\rm 1}
}
\affiliations{
    \textsuperscript{\rm 1}School of EECS, Oregon State University,
\textsuperscript{\rm 2}School of EECS, Washington State University, \\
\textsuperscript{\rm 3}Irrigated Agriculture Research and Extension Center, Washington State University

    saxenaa@oregonstate.edu, p.pesantezcabrera@wsu.edu, ballaprr@oregonstate.edu, mkeller@wsu.edu, afern@oregonstate.edu
}

\usepackage{bibentry}

\begin{document}

\maketitle

\begin{abstract}
Grapevine budbreak is a key phenological stage of seasonal development, which serves as a signal for the onset of active growth.  
This is also when grape plants are most vulnerable to damage from freezing temperatures.  
Hence, it is important for winegrowers to anticipate the day of budbreak occurrence to protect their vineyards from late spring frost events. 
This work investigates deep learning for budbreak prediction using data collected for multiple grape cultivars. While some cultivars have over 30 seasons of data others have as little as 4 seasons, which can adversely impact prediction accuracy. To address this issue, we investigate multi-task learning, which combines data across all cultivars to make predictions for individual cultivars. Our main result shows that several variants of multi-task learning are all able to significantly improve prediction accuracy compared to learning for each cultivar independently.
\end{abstract}

\section{Introduction}
In temperate climates, perennial plants such as grapevines (\textit{Vitis} spp.) undergo alternating cycles of growth and dormancy. During dormancy, the shoot and flower primordia are protected by bud scales and can reach considerable levels of cold tolerance or hardiness to maximize winter survival \cite{2020keller}. Budbreak is identified as stage 4 on the modified E-L scale \cite{coombe1995} and it is strongly influenced by the dormancy period. Once the shoots start to grow out during the process of budbreak in spring, the emerging green tissues become highly vulnerable to frost damage. 

An important issue is that ongoing climate change is increasing the risk of spring frost damage in vineyards because rising temperatures are associated with earlier budbreak and weather patterns are becoming more variable \cite{Poniajev.2022.22011}. Consequently, the ability to predict the timing of budbreak would enable producers to timely deploy frost mitigation measures (e.g. wind machines) and improve the scheduling of vineyard activities such as pruning to adjust crop load. Also, knowing when different grape varieties break bud under certain temperature scenarios enables investors and vineyard developers to better match more vulnerable varieties to lower-risk sites. 

Several models have been proposed to assess the challenging task of budbreak prediction \cite{nendel_grapevine_2010}, \cite{ferguson_modeling_2014}, \cite{zapata_predicting_2017}, \cite{camargo-a_predicting_2017}, \cite{leolini_phenological_2020}, \cite{pina-rey_phenological_2021}. As discussed by Leolini et al. these phenological models can be classified into two main categories: forcing (\emph{F}) and chilling-forcing (\emph{CF}) models. On one hand, forcing models are based on the accumulation of forcing units from a fixed date in the year. \emph{F} models focus solely on describing the eco-dormancy period by assuming that the endo-dormancy period has ended and the chilling unit accumulation requirement has been met. On the other hand, \emph{CF} models account for both the endo- and eco-dormancy periods by considering the chilling unit and the forcing accumulation in relation to specific temperature thresholds ---i.e., an estimated base temperature $T_b$. Although these models take into account thermal requirements, none of them include other environmental variables (e.g., solar radiation, relative humidity, precipitation, dew point) besides air temperature. 

The aim of this study is to investigate modern deep learning techniques for incorporating a wider range of weather data into budbreak predictions. In particular, we develop a Recurrent Neural Network (RNN) for budbreak prediction from time series input of various weather features. The proposed models' performance tends to degrade in the case of cultivars that have limited data. Multi-Task Learning has the potential to alleviate this issue, as it can utilize data across all cultivars to improve budbreak prediction. The main contributions of this work are: 1) to frame this multi-cultivar learning problem as an instance of multi-task learning, and 2) to propose and evaluate a variety of multi-task RNN models on real-world data. Finally, the obtained results show that multi-task learning is able to significantly outperform single-task learning. Due to lack of programmatic access to existing budbreak models at the time of this writing, we reserve a comparison to those models for future work. 


\section{Datasets}
This study used phenological data collected for 31 diverse grape cultivars from 1988 to 2022 by the Viticulture Program at WSU Irrigated Agriculture Research and Extension Center (IAREC). Data collection was performed in the vineyards of the IAREC, Prosser, WA (46.29°N latitude; -119.74°W longitude) and the WSU-Roza Research Farm, Prosser, WA (46.25°N latitude; -119.73°W longitude). In north-south-oriented rows, the vineyards were planted in a fine sandy loam soil type with vine spacing of 2.7m between rows and 1.8m within rows. A regulated deficit irrigation system was used to drip-irrigate the vines, and they were spur-pruned and trained to a bilateral cordon \cite{zapata_predicting_2017}. 

Phenological data were collected as the day of year (DOY) when a particular phenological stage, ranging from bud first swell to harvest, was observed. The budbreak stage is defined as the presence of green tissue in 50\% of previously dormant buds \cite{ferguson_modeling_2014, zapata_predicting_2017}. Additionally, the API provided by AgWeatherNet was used to obtain environmental daily data from the closest on-site weather station to each cultivar \cite{AgWeatherNet}. The two stations used are Prosser.NE (46.25°N latitude; -119.74°W longitude) and Roza.2 (46.25°N latitude; -119.73°W longitude).
Thus, a continuously growing dataset containing a variable number of years of daily weather data is created for each cultivar, along with phenological stage labels placed in the corresponding DOY when observed. 

Table 1 shows a summary of the number of years of data collected for the different cultivars. The interval of years in parenthesis represents the years with no phenological data.

\begin{table}[htb]
\centering
\resizebox{1\columnwidth}{!}{
\begin{tabular}{ |l |l |c|c| }
\hline
\multicolumn{1}{|c|}{\textbf{Cultivar}}             &\multicolumn{1}{|c|}{\textbf{Phenology Years}}                      &\makecell{\textbf{Phenology Total} \\ \textbf{Years of Data}} \\\hline
Barbera	             &2015-2022	                                    &8                          \\\hline
Cabernet Sauvignon	 &1988-2022 (1999, 2007, 2008, 2012-2014)	    &29                         \\\hline
Chardonnay	         &1988-2022 (1989, 1996, 1999, 2011-2014)	    &28                         \\\hline
Chenin Blanc	     &1988-2022 (1996, 2007-2014)	                &26                         \\\hline
Concord	             &1992-2022 (2011-2014)	                        &27                         \\\hline
Grenache	         &1992-2022 (2007-2014)	                        &23                         \\\hline
Malbec	             &1988-2022 (1996, 2004-2014)	                &23                         \\\hline
Merlot	             &1988-2022 (1996, 2011-2014)	                &30                         \\\hline
Mourvedre	         &2015-2022 	                                &8                          \\\hline
Nebbiolo	         &2015-2022	                                    &8                          \\\hline
Pinot Gris	         &1992-2022 (2007-2014)	                        &23                         \\\hline
Riesling	         &1988-2022 (1996, 2008, 2009, 2011-2014)	    &28                         \\\hline
Sangiovese	         &2015-2022                                     &8                          \\\hline
Sauvignon Blanc	     &2004-2022 (2007-2014)	                        &11                         \\\hline
Semillon	         &1988-2022 (1996, 2007-2014)                   &26                         \\\hline
Syrah	             &2015-2018	                                    &4                          \\\hline
Viognier	         &2015-2022	                                    &8                          \\\hline
Zinfandel	         &1992-2022 (1996, 2007-2014)	                &22                         \\\hline
\end{tabular}
}

\caption{Summary of phenology data collection of selected cultivars.}
\label{tab:data-description}
\end{table}

\begin{figure*}[t]
\centering
\includegraphics[width=0.97\textwidth]{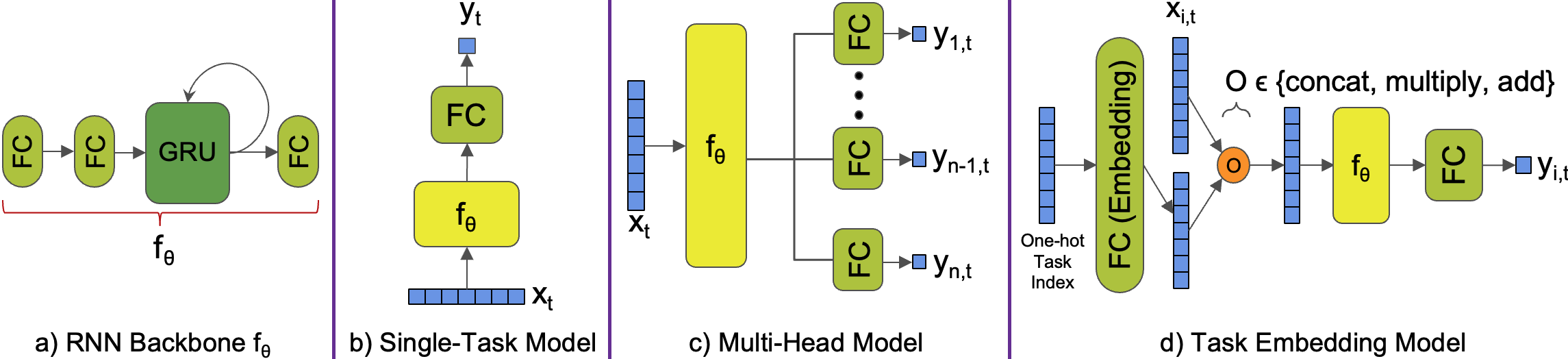} 
\caption{Network Architectures. FC denotes fully connected layers and GRU denotes Gated Recurrent Unit. a) The RNN backbone processes data sequences ($X_t$). b) The STL model with a single prediction layer. c) Multi-Head MTL variant which has a prediction layer per cultivar. d) Task Embedding MTL variant, which considers the task at hand as an input}
\label{fig:model-diagrams}
\end{figure*}

\section{Budbreak Prediction Models and Training}
We formulate budbreak prediction as a sequence prediction problem. 
We represent the sequential data for cultivar $i$ in year $k$ by $S_{i,k} = (x_1, y_1, x_2, y_2, \ldots, x_H, y_H)$, where $H$ is the number of days in the year (accounting for leap year), $x_t$ represents the weather data, and $y_t$ represents the ground truth budbreak label for the day $t$. The label $y_t$ is 1 if budbreak occurred before or at day $t$ and is 0 otherwise. Thus, $y_t$ is a step function that rises from 0 to 1 on the day of budbreak.  
A dataset for cultivar $i$ is denoted by $D_i = \{S_{i,k} \;|\; k\in \{1,\ldots,N_i\}\}$, where $N_i$ is the number of seasons with data for cultivar $i$. Based on these datasets, our goal is to learn a model $M_i$ for each cultivar that takes in weather features up to any day $t$ and outputs a probability of budbreak for the day $t$. Note that in practice such a model can be used for making budbreak projections into the future by feeding the model with weather forecasts. 


The most common learning paradigm is single-task learning (STL), which for our problem corresponds to learning a cultivar model $M_i$ from only that cultivar's data $D_i$. This paradigm can work well when enough data is available for a cultivar. However, for low-data cultivars (e.g. 4 seasons) we can expect prediction accuracy to suffer. 
To address this issue, we consider a multi-task learning (MTL) paradigm, which uses data across all cultivars to make predictions for individual cultivars. Assuming that different cultivars share common budbreak characteristics, this approach has the potential to improve accuracy over STL. Below we describe the deep-learning-based STL and MTL models that we use in this work.



\subsection{Single-Task Model}
Our STL model makes causal budbreak predictions by sequentially processing a weather data sequence $x_1,x_2,\ldots, x_t$ and at each step outputting the corresponding budbreak probability estimate.  For this purpose, we use a recurrent neural network (RNN) \cite{RNN}, which is a widely used model for sequence data. The RNN backbone used by both our STL and MTL models is illustrated in Figure \ref{fig:model-diagrams}a, which we denote by $f_{\theta}$ with parameters $\theta$. The backbone network begins with two fully connected (FC) layers, followed by a gated recurrent unit (GRU) layer \cite{GRU}, which is followed by another FC layer. 

Our STL model, shown in Figure \ref{fig:model-diagrams}b, simply feeds daily weather data $x_t$ into the first FC layer as input and adds an additional FC layer to produce the final LTE prediction output. Intuitively, the GRU unit, through its recurrent connection is able to build a latent-state representation of the sequence data that has been processed so far. For our budbreak problem, this representation should capture information about the weather history which is useful for predicting budreak. In some sense, the latent state can be thought of as implicitly approximating the internal state of the plant as it evolves during the year. As described below, each STL model $M_i$ is trained independently on its cultivar-specific dataset $D_i$.

\subsection{Multi-Task Models}
We consider two types of MTL models that directly extend the RNN backbone of Figure \ref{fig:model-diagrams}a, the multi-head model and the task-embedding model.

{\bf Multi-Head Model.} The multi-head model is perhaps the most straightforward approach to MTL and has been quite successful in prior work when tasks are highly related \cite{mtlfirstpaper}. As illustrated in Figure \ref{fig:model-diagrams}c, the multi-head model is identical to the STL model, except, that it adds $C$ parallel cultivar-specific fully-connected layers to the backbone (i.e. prediction heads). Each prediction head is responsible for producing the budbreak prediction for its designated cultivar. This model allows the cultivars to share the features produced by the RNN backbone, with each cultivar-specific output simply being a linear combination of the shared features. Intuitively, if there are common underlying features that are useful across cultivars, then this architecture allows those to emerge based on the combined set of data. Thus, cultivars with small amounts of data can leverage those useful features and simply need to tune a set of linear weights based on the available data. We abbreviate this model as \textit{MultiH} in future sections.

{\bf Task-Embedding Models.} Our task embedding model for MTL is similar in spirit to prior work \cite{csnn,taskembedding1,taskembedding2} and motivated by the form of typical scientific models. Scientists define the overall mechanisms and structure of those models along with a fixed set of model parameters that can be tuned for specific cultivars (e.g. chill accumulation rate). Similarly, the task embedding model uses a neural network to learn a general model that accepts cultivar-specific parameters as well as learning the parameters for each cultivar. 

As illustrated in Figure \ref{fig:model-diagrams}d, the task embedding model first maps a one-hot encoding of the cultivar in consideration to an embedding vector (analogous to cultivar ``parameters"), which is combined with the weather data $x_t$ and then fed to the GRU unit. This allows for predictions to be specialized for each cultivar. Intuitively cultivars with more similar budbreak characteristics will have more similar embedding vectors. 
We explore three variants of this architecture that differ in how they combine the embedding with the weather data: \textit{AddE} simply adds the vectors together, \textit{ConcatE} simply concatenates the vectors, and \textit{MultE} does element-wise multiplication of the vectors. 

\subsection{Model and Training Details}


Our models use the following daily weather features that capture: \textit{Temperature, Humidity, Dew Point, Precipitation, and Wind Speed}.
We handle missing weather data via linear interpolation. 

We run three training trials for each of our models and report averages across tries. Each trial selects 2 different seasons to use as test data for reporting performance and uses the remaining data for training. The models are trained to minimize the binary cross entropy (BCE) loss between the predicted budbreak probability at each step and the true budbreak label. We use Adam \cite{adam} as the optimizer with a learning rate of 0.001 and a batch size of 12 seasons shuffled randomly. We train all our models for 400 epochs. The output dimensionality of the linear layers of the RNN backbone are 1024, 2048, and 1024 respectively. The GRU has a hidden state and internal memory of dimensionality 2048.

\section{Experiments}
Our experiments involve 18 cultivars with amounts of data ranging from 4 to 23 years. 
\begin{table}[htb]
\centering
\resizebox{1\columnwidth}{!}{
\begin{tabular}{ |l |r|r|r|r| }
\hline
\multicolumn{1}{|c|}{\textbf{Cultivar}}	&	\textbf{MultE}	&	\textbf{ConcatE}	&	\textbf{AddE}	&	\textbf{MultiH}\\\hline
Barbera	&	1.69	&	1.91	&	1.85	&	1.91\\\hline
Cabernet Sauvignon	&	-0.05	&	-0.05	&	-0.05	&	-0.07\\\hline
Chardonnay	&	-0.20	&	1.15	&	1.39	&	-0.15\\\hline
Chenin Blanc	&	0.06	&	0.01	&	-0.20	&	-0.11\\\hline
Concord	&	0.06	&	0.02	&	-0.01	&	0.02\\\hline
Grenache	&	1.23	&	1.20	&	1.19	&	1.20\\\hline
Malbec	&	3.85	&	3.92	&	3.86	&	3.87\\\hline
Merlot	&	-0.96	&	0.29	&	-0.01	&	-0.17\\\hline
Mourvedre	&	9.56	&	10.04	&	10.03	&	10.09\\\hline
Nebbiolo	&	0.82	&	0.93	&	0.90	&	0.86\\\hline
Pinot Gris	&	-0.24	&	-0.02	&	-0.03	&	-0.04\\\hline
Riesling	&	0.03	&	0.02	&	-0.07	&	-0.02\\\hline
Sangiovese	&	18.38	&	18.55	&	18.61	&	18.57\\\hline
Sauvignon Blanc	&	-0.02	&	0.17	&	0.14	&	0.16\\\hline
Semillon	&	0.13	&	0.14	&	0.07	&	0.13\\\hline
Syrah	&	3.69	&	3.88	&	3.79	&	3.86\\\hline
Viognier	&	-0.17	&	0.38	&	0.43	&	0.39\\\hline
Zinfandel	&	-0.15	&	0.11	&	0.11	&	0.15\\\hline
\end{tabular}
}
\caption{Difference in BCE between MTL models and the basline STL model for each cultivar. Positive values indicates MTL improves over STL.}
\label{tab:results}
\end{table}

\textbf{Single-Task Learning vs Multi-Task Learning.}

Table \ref{tab:results} shows the difference in BCE between our MTL models and the baseline STL model for each of the cultivars. A positive value indicates that the MTL model improved over the STL model in terms of BCE. We observe that for most cultivars all of the MTL variants improve over STL. For some cultivars, there are very large improvements, e.g Sangiovese and Syrah. The different MTL variants typically perform similarly.
However, we see that if we consider the number of cultivars where MTL slightly underperforms STL, ConcatE appears to have a slight advantage as it only underperforms on two cultivars.

\begin{figure}[tbh]
\centering
\includegraphics[width=1\columnwidth]{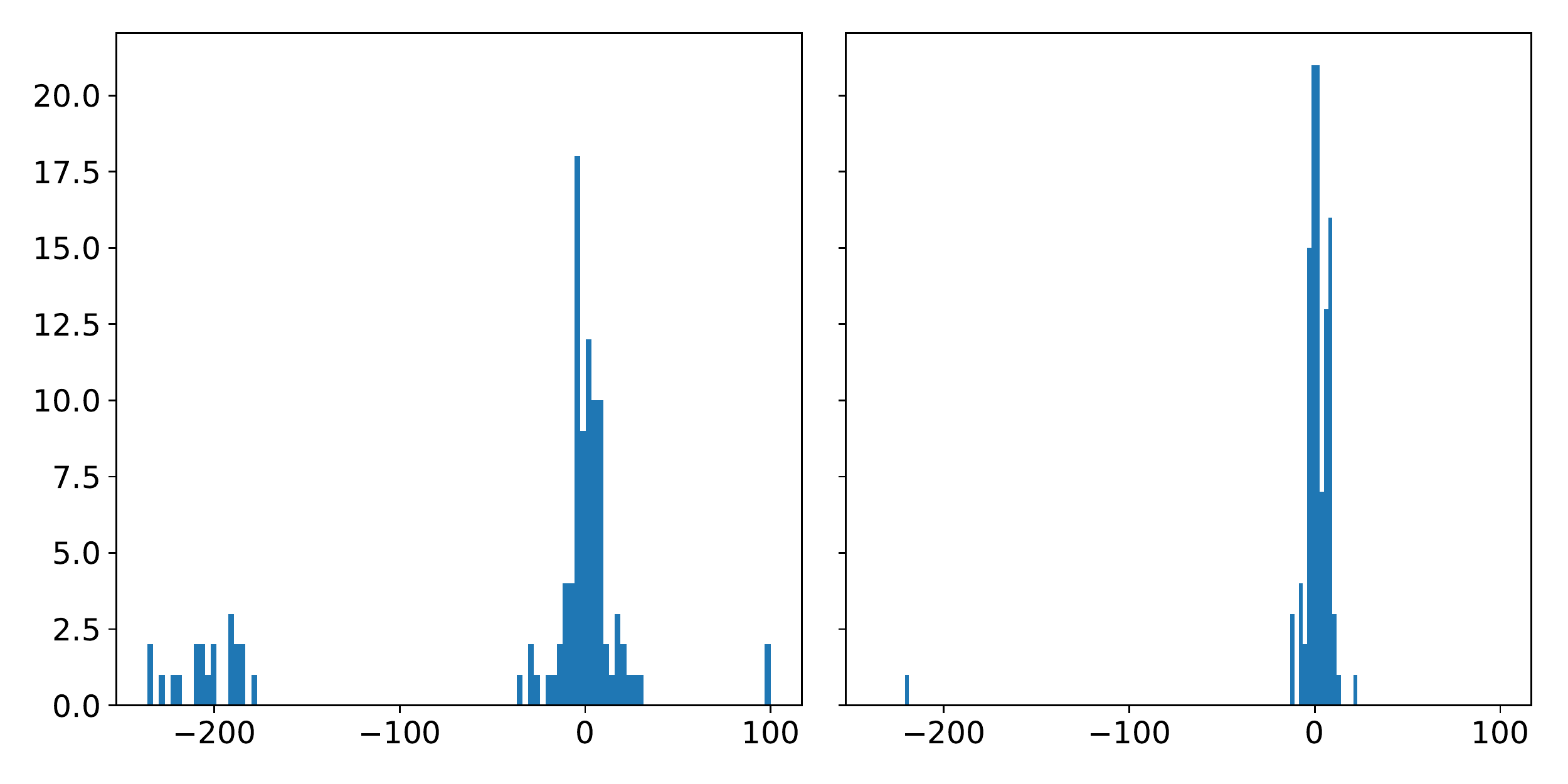} 
\caption{Comparison of Histograms (STL on the left and MultiHead MTL on the right) of the difference in days metric. Observe that the STL histogram has more outliers than the MultiHead histogram. The x-axis denotes the difference in day metric and the y-axis denotes the frequency of occurrence of that metric.}
\label{fig:hist}
\end{figure}

\begin{table}[tbh]
\centering
\resizebox{\columnwidth}{!}{
\begin{tabular}{|l|c|c|l|l|l|}
\hline
Model & Median & \textgreater{}3days & \textgreater{}1week & \textgreater{}2weeks & \textgreater{}1month \\ \hline
Single  & 7   & 32 & 19 & 13 & 25 \\ \hline
AddE    & 3   & 29 & 21 & 5  & 6  \\ \hline
MultE   & 5   & 38 & 25 & 10 & 8  \\ \hline
ConcatE & 3.5 & 34 & 24 & 1  & 1  \\ \hline
MultiH  & 3   & 28 & 30 & 1  & 1  \\ \hline
\end{tabular}%
}
\caption{Looking at the difference of days metric for different model variants. We see that all the multi-task learning variants improve over STL.}
\label{tab:datasetsize}
\end{table}
\textbf{Difference in days metric.}
To get a better understanding of the practical differences between MTL and STL we now consider using the models to predict the day of budbreak. In particular, we use a simple approach of predicting budbreak starting on the first day when the predicted probability is more than 0.5. Figure \ref{fig:hist} shows two histograms of the differences between the predicted budbreak day and the ground truth day over all seasons and cultivars. The first histogram is for the MTL MultiH model and the second is for STL. Results are similar for other MTL models. We see that there are many more outliers predictions with large errors for the STL model compared to the MTL model. Table \ref{tab:datasetsize} breaks down these results further and shows the median absolute error in day prediction along with the number of predictions that fall beyond selected error thresholds. We see that the medians for MTL models are significantly better than for STL. Further, the MTL ConcatE and MultiH models produce the fewest larger errors of two weeks or more. 

\begin{figure}[tbp]
\centering
\subfloat[STL CE Loss 0.766 difference in days -201]{
	\label{subfig:correct}
	\includegraphics[width=0.45\textwidth]{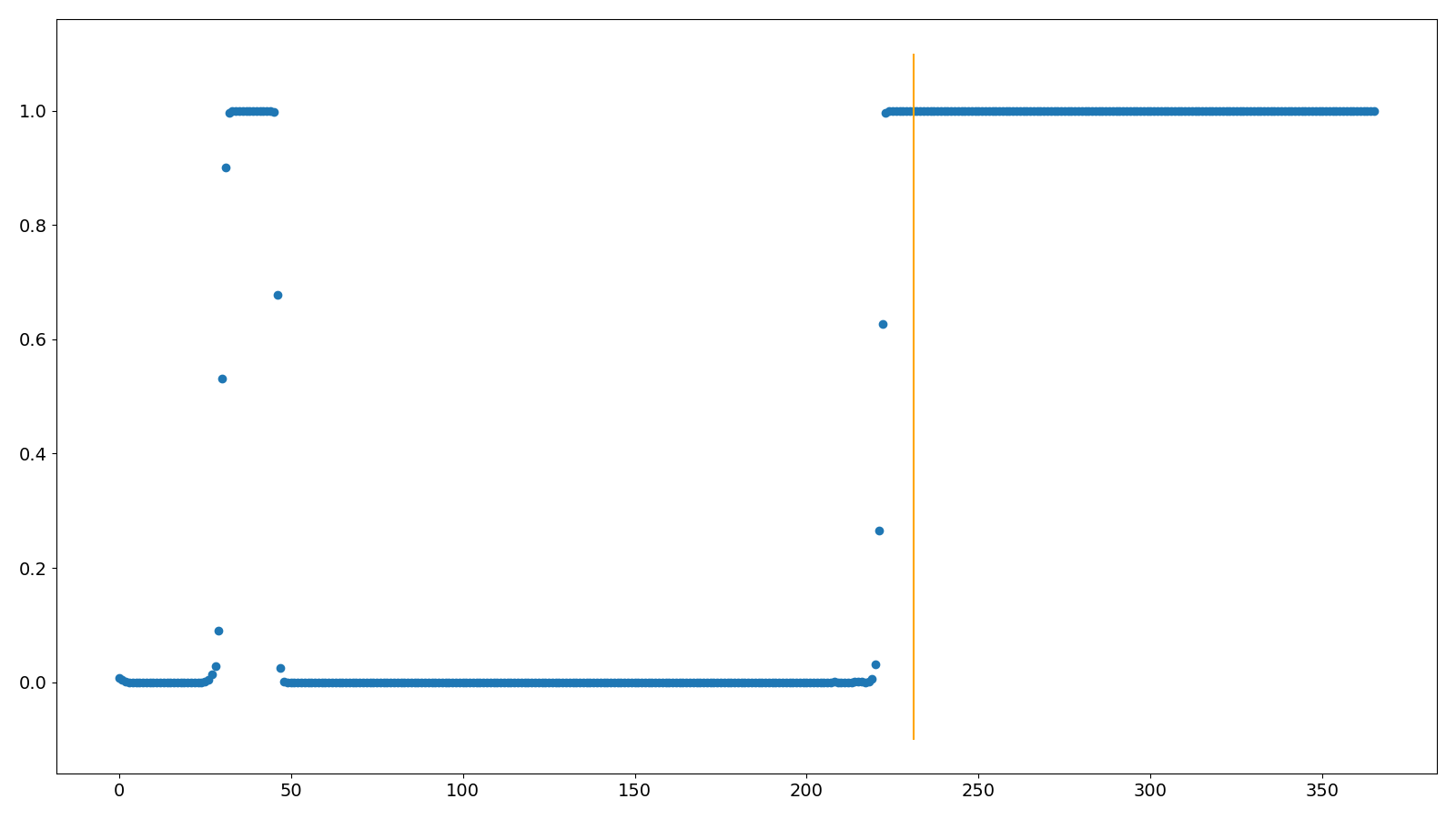}} 

\subfloat[MTL CE Loss 0.039 difference in days 4]{
	\label{subfig:notwhitelight}
	\includegraphics[width=0.45\textwidth]{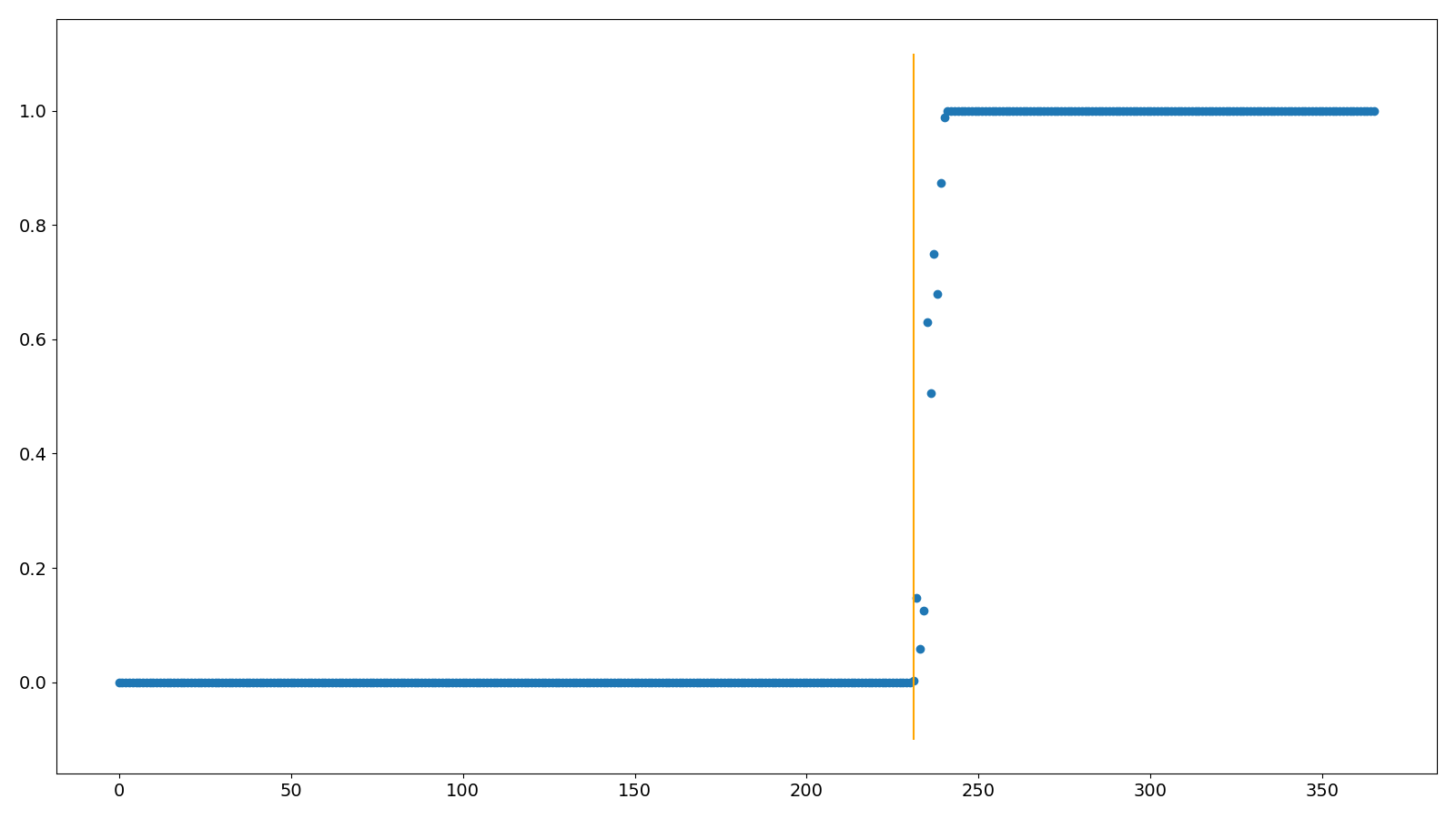} } 
\caption{Comparing Budbreak prediction for the STL and MTL (MultiHead) models for the Syrah cultivar. Note that the STL model is unable to predict the correct shape of the function (step function). The x-axis denotes the day of the year and the y-axis indicates the probability of budbreak.}
\label{fig:bars}
\end{figure}

To get insight into the nature of the large errors in STL compared to MTL, Figure \ref{fig:bars} shows the predicted probabilities for the STL and MultiH model for a particular cultivar and season where a large STL error occurred. We see that the STL model produced a very early jump in probability, possibly resulting from an unusually warm time period. Rather, the MTL model avoids the early jump in probability, which is likely due to learning a better general model of budbreak characteristics based on the larger amount of data available from other cultivars.


\section{Conclusion}
This study showed the effectiveness of multi-task learning for budbreak prediction. In the immediate future, we wish to incorporate more phenological stages in our budbreak prediction model. Curated models will be deployed on AgWeatherNet in the 2023-2024 season. Subsequent work will focus on investigating the utility of MTL for other agriculture-related problems with limited data.

\section*{Acknowledgements}
This research was supported by USDA NIFA award No. 2021-67021-35344 (AgAID AI Institute).
The authors thank Lynn Mills for the collection of phenological data.

\bibliography{aaai23}

\begin{thebibliography}{17}
\providecommand{\natexlab}[1]{#1}

\bibitem[{Camargo-A. et~al.(2017)Camargo-A., Salazar-G., Zapata, and
  Hoogenboom}]{camargo-a_predicting_2017}
Camargo-A., H.; Salazar-G., M.; Zapata, D.; and Hoogenboom, G. 2017.
\newblock Predicting the dormancy and bud break dates for grapevines.
\newblock \emph{Acta Horticulturae}, (1182): 153--160.

\bibitem[{Caruana(1997)}]{mtlfirstpaper}
Caruana, R. 1997.
\newblock Multitask learning.
\newblock \emph{Machine learning}, 28(1): 41--75.

\bibitem[{Cho et~al.(2014)Cho, van Merrienboer, Bahdanau, and Bengio}]{GRU}
Cho, K.; van Merrienboer, B.; Bahdanau, D.; and Bengio, Y. 2014.
\newblock On the Properties of Neural Machine Translation: Encoder-Decoder
  Approaches.

\bibitem[{Coombe(1995)}]{coombe1995}
Coombe, B. 1995.
\newblock Growth Stages of the Grapevine: Adoption of a system for identifying
  grapevine growth stages.
\newblock \emph{Australian Journal of Grape and Wine Research}, 1(2): 104--110.

\bibitem[{Ferguson et~al.(2014)Ferguson, Moyer, Mills, Hoogenboom, and
  Keller}]{ferguson_modeling_2014}
Ferguson, J.~C.; Moyer, M.~M.; Mills, L.~J.; Hoogenboom, G.; and Keller, M.
  2014.
\newblock Modeling {Dormant} {Bud} {Cold} {Hardiness} and {Budbreak} in
  {Twenty}-{Three} {Vitis} {Genotypes} {Reveals} {Variation} by {Region} of
  {Origin}.
\newblock \emph{American Journal of Enology and Viticulture}, 65(1): 59--71.

\bibitem[{Keller(2020)}]{2020keller}
Keller, M., ed. 2020.
\newblock \emph{The Science of Grapevines}.
\newblock London: Elsevier Academic Press, third edition.
\newblock ISBN 978-0-12-816365-8.

\bibitem[{Kingma and Ba(2014)}]{adam}
Kingma, D.~P.; and Ba, J. 2014.
\newblock Adam: A Method for Stochastic Optimization.

\bibitem[{Leolini et~al.(2020)Leolini, Costafreda-Aumedes, A.~Santos, Menz,
  Fraga, Molitor, Merante, Junk, Kartschall, Destrac-Irvine, van Leeuwen,
  C.~Malheiro, Eiras-Dias, Silvestre, Dibari, Bindi, and
  Moriondo}]{leolini_phenological_2020}
Leolini, L.; Costafreda-Aumedes, S.; A.~Santos, J.; Menz, C.; Fraga, H.;
  Molitor, D.; Merante, P.; Junk, J.; Kartschall, T.; Destrac-Irvine, A.; van
  Leeuwen, C.; C.~Malheiro, A.; Eiras-Dias, J.; Silvestre, J.; Dibari, C.;
  Bindi, M.; and Moriondo, M. 2020.
\newblock Phenological {Model} {Intercomparison} for {Estimating} {Grapevine}
  {Budbreak} {Date} ({Vitis} vinifera {L}.) in {Europe}.
\newblock \emph{Applied Sciences}, 10(11): 3800.
\newblock Number: 11 Publisher: Multidisciplinary Digital Publishing Institute.

\bibitem[{Nendel(2010)}]{nendel_grapevine_2010}
Nendel, C. 2010.
\newblock Grapevine bud break prediction for cool winter climates.
\newblock \emph{International Journal of Biometeorology}, 54(3): 231--241.

\bibitem[{Piña-Rey et~al.(2021)Piña-Rey, Ribeiro, Fernández-González,
  Abreu, and Rodríguez-Rajo}]{pina-rey_phenological_2021}
Piña-Rey, A.; Ribeiro, H.; Fernández-González, M.; Abreu, I.; and
  Rodríguez-Rajo, F.~J. 2021.
\newblock Phenological {Model} to {Predict} {Budbreak} and {Flowering} {Dates}
  of {Four} {Vitis} vinifera {L}. {Cultivars} {Cultivated} in {DO}. {Ribeiro}
  ({North}-{West} {Spain}).
\newblock \emph{Plants}, 10(3): 502.

\bibitem[{Poni, Sabbatini, and Palliotti(2022)}]{Poniajev.2022.22011}
Poni, S.; Sabbatini, P.; and Palliotti, A. 2022.
\newblock Facing Spring Frost Damage in Grapevine: Recent Developments and the
  Role of Delayed Winter Pruning {\textendash} A Review.
\newblock \emph{American Journal of Enology and Viticulture}.

\bibitem[{Rumelhart, Hinton, and Williams(1985)}]{RNN}
Rumelhart, D.~E.; Hinton, G.~E.; and Williams, R.~J. 1985.
\newblock Learning internal representations by error propagation.
\newblock Technical report, California Univ San Diego La Jolla Inst for
  Cognitive Science.

\bibitem[{Schreiber and Sick(2021)}]{taskembedding2}
Schreiber, J.; and Sick, B. 2021.
\newblock Emerging relation network and task embedding for multi-task
  regression problems.
\newblock In \emph{2020 25th International Conference on Pattern Recognition
  (ICPR)}, 2663--2670. IEEE.

\bibitem[{Schreiber, Vogt, and Sick(2021)}]{taskembedding1}
Schreiber, J.; Vogt, S.; and Sick, B. 2021.
\newblock Task Embedding Temporal Convolution Networks for Transfer Learning
  Problems in Renewable Power Time Series Forecast.
\newblock In \emph{Joint European Conference on Machine Learning and Knowledge
  Discovery in Databases}, 118--134. Springer.

\bibitem[{Silver, Poirier, and Currie(2008)}]{csnn}
Silver, D.~L.; Poirier, R.; and Currie, D. 2008.
\newblock Inductive transfer with context-sensitive neural networks.
\newblock \emph{Machine Learning}, 73(3): 313--336.

\bibitem[{WSU(2022)}]{AgWeatherNet}
WSU. 2022.
\newblock AgWeatherNet {\textbar} Daily Data.

\bibitem[{Zapata et~al.(2017)Zapata, Salazar-Gutierrez, Chaves, Keller, and
  Hoogenboom}]{zapata_predicting_2017}
Zapata, D.; Salazar-Gutierrez, M.; Chaves, B.; Keller, M.; and Hoogenboom, G.
  2017.
\newblock Predicting {Key} {Phenological} {Stages} for 17 {Grapevine}
  {Cultivars} ({Vitis} vinifera {L}.).
\newblock \emph{American Journal of Enology and Viticulture}, 68(1): 60--72.
\newblock Publisher: American Journal of Enology and Viticulture Section:
  Research Article.

\end{thebibliography}
\end{document}